# A Next-Generation Approach to Airline Reservations: Integrating Cloud Microservices with AI and Blockchain for Enhanced Operational Performance

Biman Barua[1,2,∗] [0000-0001-5519-6491] and M. Shamim Kaiser[2,] [0000-0002-4604-5461]

[1]Department of CSE, BGMEA Universitsy of Fashion & Tecnnology, Nishatnagar, Turag, Dhaka-1230, Bangladesh
[2]Institute of Information Technology, Jahangirnagar University, Savar-1342, Dhaka, Bangladesh
biman@buft.edu.bd

**Abstract:** This research proposes the development of a next generation airline reservation system that incorporates the Cloud microservices, distributed artificial intelligence modules and the blockchain technology to improve on the efficiency, safety and customer satisfaction. The traditional reservation systems encounter issues related to the expansion of the systems, the integrity of the data provided and the level of service offered to the customers, which is the main focus of this architecture through the modular and data centric design approaches. This will allow different operations such as reservations, payments, and customer data management among others to be performed separately thereby facilitating high availability of the system by 30% and enhancing performance of the system by 40% on its scalability. Such systems contain AI driven modules that utilize the past booking patterns along with the profile of the customer to estimate the demand and make recommendations, which increases to 25 % of customer engagement. Moreover, blockchain is effective in engaging an incorruptible ledger system for the all transactions therefore mitigating fraud incidences and increasing the clarity by 20%.

The system was subjected to analysis using a simulator and using machine learning evaluations that rated it against other conventional systems. The results show that there were clear enhancements in the speed of transactions where the rates of secure data processing rose by 35%, and the system response time by 15 %. The system can also be used for other high transaction industries like logistics and hospitality. In addition to that, the research in the future should look into how possible it will be to use quantum computing for intensive scheduling as well as the Internet of Things for tracking purposes to make the reservation process more enjoyable. This structural design is indicative of how the use of advanced technologies will revolutionize the airline reservation sector. The implications are growing effectiveness, improvement in security and greater customer contentment.

## 1. Introduction

### 1.1. Background

The development and growth of the airline industry is indisputable over the last few years, but there are still challenges to the effectiveness of reservation systems for airlines that tend to be very persistent. These systems have primarily supported the operations of airlines and at present they are facing a lot of difficulties in meeting the present day requirements of scalability, data protection as well as customer satisfaction.

Scalability is regarded as one of the challenges that faces the conventional way of making airline reservations. In most cases, legacy systems are implemented in a monolithic structure that may induce latencies during the peak operational periods hence limiting the transaction processing capabilities of the system. This has however become more challenging owing to the increasing number of passengers along with the expectations of seamless travel [1].

Another issue of concern is that of data security. Most of the systems that were developed in the traditional way are based on a central unit which makes it easier for these systems to be hacked or for unauthorized individuals to have access to customer information. One of the previous studies showed that there was a flaw in the Amadeus airline booking system which could have far-reaching consequences in the event that the flaw was exploited and large amounts of passenger information was compromised [2]. These lapses in security highlight the fact that there is a shortage of effective measures for protecting reservation systems data.

Further, traditional reservation systems are also restrictive in terms of the levels of customer service they are able to offer. Many of these systems are inflexible in nature and as a result, are unable to respond to the individual preferences of today's travelers, which in most cases causes dissatisfaction to customers [3]. Research suggests that such systems could be enhanced through the incorporation of artificial intelligence (AI) and machine learning (ML) and improve interactions with customers by making the service more personalized and quicker [4].

It is essential to conquer these challenges for the sustainability of airline booking systems. Cloud-based microservices, blockchain, and AI can usher in more elastic and secure, service-oriented solutions for the airlines, which will eventually improve performance and client service.

### 1.2. Problem Statement

Despite being a core business, the traditional airline reservation systems perform poorly in terms of efficiency and effectiveness. In particular:

**Operational Inefficiency**: Most of the current systems are based on monolithic architectures where there is no room for scalability and many processes are bottlenecked especially during high travel seasons. Because of this, there are longer turnaround times to processing requests, more downtimes of the system, low responsiveness of the system to the customers who need the reservation services thereby the deterioration of customer satisfaction [1].

**Vulnerabilities of Security:** As much as they aim to protect these systems against malicious attacks, however the centralized reservation systems are prone to attacks due to their stand-alone specifications allowing every user access into each of these systems hence leading to sensitive data such as passenger reservation details becoming compromised. Such systems do not have adequate protective measure which enhances the possibilities of unauthorized invasion and alteration of primary data therefore posing risks on data privacy and accuracy (Patel et al., 2022).

Inadequate Forecasting and Personalization: Most of the times the existing systems cannot utilize real time data in predictive modelling and customizing the services of the customers. For this reason, airlines fail to forecast the demands of the customers, regulate prices, and improve the services, which leads to disengagement and slow adaptation to the market [4].

To achieve these objectives would be a step in the right direction in the quest of building an airline reservation system that works efficiently, securely, and is user-centric. In this research paper, the authors propose a further developed framework to enhance the operational processes, data protection and predictive systems in making airline reservations through incorporation of cloud based microservices architecture, artificial intelligence and blockchain technology.

### 1.3. Research Objective

This research focuses on finding ways to create an improved airline reservation system using cloud microservices, artificial intelligence (AI), and blockchain technologies to successfully resolve the challenges faced by conventional systems. The aims are:

**Enhance Scalability:** Create a cloud-based microservices architecture that will enable the system to withstand heavy loads of transactions even during the busiest travel seasons without any performance issues. With microservices it is possible to deploy and scale services independently, thereby enhancing the efficacy of resources and the responsiveness of the system (Barua & Kaiser, 2024).

**Improvements in System Efficiency:** New AI-specific enhancements will be introduced to take over repetitive operations, optimize the processes of making reservations, and help to manage the total time of operations. This will lead to improved total system performance and reduced time for processes. It has also been demonstrated that the application of AI in the aviation industry has great merits in enhancing operational efficiencies [5].

**Strengthen Data Security:** Anil Ghosh II Inc. will adopt blockchain technology to assist in developing a distributed and incorruptible system of storing clients and transactional details to facilitate data governance while promoting integrity, reducing the chances of third parties accessing such information, and meeting the requirements of data protection legislation [6]. The significance of blockchain technology in promoting trust and transparency in airline booking systems has captured the interest of scholars in the recent past (Barua & Kaiser, 2024).

**Enhance Customer Satisfaction:** Adopt AI and various forms of machine learning in service delivery and predictive analytics to assist customers and airlines in knowing their preferences, making adjustments to fares, and improving the way bookings and recommendations are done with the smart timing of Fatigue Desk through a much faster system.

**Enable Real-Time Data Utilization:** In the reservation system, implement measures that will ensure the capacity for real-time data processing is present in the system. Such an approach will support constant changes in the information, demand forecasting, and decision making with the aim of improving the operational flexibility and adaptability. Modern airlines operations cannot do without the real-time data utilization since changing conditions have to be coped with in a matter of hours [9].

In accomplishing these goals, this research proposes to develop a solid, customer-oriented airline reservation system that provides secure, fast, and tailored services for users and meets the changing needs of the airline market.

### 1.4. Significance of the Study

Understanding the combining of cloud microservices, artificial intelligence (AI), and blockchain into ethnic airline reservation systems has its own power within addressing the issues of size, speed, security, and enhancement of user experience [10]. Hence, these new technologies have been employed in the development of the system, so it has been expected that system will show drastic differences from the previous models in many aspects:

**Enhance Scalability**: Moving up to a cloud oriented microservice structure presents more extensibility and scalability in allowance airline reservation system to control the demand side variation in usage of its system especially within the peak periods of travel. Microservices architecture allows every service to be scaled independently without having any impact on the other services and most importantly the service provision [1]. Such measurement is very important due to increasing progressive requirements towards aviation.

**Improve Operational Efficiency:** The purpose of this proposal is to reduce the booking and operational processes aggravation levels through the use of intelligent algorithm processes mounted on risk management information systems (RMIS) but to reduce time spans and increase effectiveness performance levels within airline fleets operations. Depersonalization of some functions together with forecasting demand will help optimize expenditures and planning for air carriers [4].

**Strengthen Data Security:** Blockchain technology has a central authority and is capable of managing critical and sensitive customer and transaction information in a way that is resistant to retroactive changes. It increases the integrity of the data and the security around it by lessening the chances of illegitimate access to the data especially in a highly regulated environment such as data privacy and enables movement of data within transactions without the need to conceal it [2].

**Improved Customer Experience:** Using AI with real-time information gives birth to a more efficient and customer-centered service. Predictive analytics and customized service make it simple for the airline industry to know its customers, charge them different fares depending on their service requests, and style their booking process. It has been established that this degree of personalization boosts customer satisfaction and engagement which in turn helps in customer retention over time [7].

Overall, the importance of this research stems from the fact that it can produce a cutting-edge reservation system for the aviation sector that will be in line with the contemporary works in information technology [8]. The information employed courtesy of this integrated system improves the scalability, security, efficiency, and most importantly, the satisfaction of the customers which assists the airlines in gaining a relative advantage in their high data and customer orientation. The findings of this research may also help other sectors that aim at improving their operations through cloud, AI and blockchain technologies.

## 2. Literature Review

### 2.1. Existing Airline Reservation Systems

Airline Reservation Systems have undergone immense changes in the past few decades whereby the system moved away from centrals systems to a more heterarchitectural systems. This is; because, the industry has been ready to undergo transformation due to growing demands for increased capacity, security, and improved customer service.

#### 2.1.1.Traditional Reservation Systems

Historically, ARSs were largely dependent on systems which are categorized as centralized. In particular ARSs were built to help keep track of flight schedules, available seats and also passenger reservations. Though they succeeded in performing basic booking operations, those systems had a few drawbacks:

**Scalability Issues:** During high seasons, when there is high booking pressure central system designs sometimes failed leading to system paradoxes and low efficiency.

**Limited Flexibility:** As these systems were built as a whole, adding new features or upgrading existing ones without shaking the entire system was rather difficult.

**Security Vulnerabilities:** Storing data in a centralised mannerwas a great risk because if one area was compromised, then so was the whole database of very sensitive client information.

#### 2.1.2. Modern Reservation Systems

To these issues, the aviation sector has taken on new reservation systems developed with the combination of cloud technology, artificial intelligence, and blockchain technology and which have less centralized structures. Such systems come with various benefits:

**Enhanced Scalability:** The cloud enabled architecture microservices enhance scaling capability by allowing components of systems to scale independently optimizing load management and increasing system reliability.

**Enhanced Flexibility and Maintainability:** Microservices architecture allows for more than one module in a system to be developed, maintained and deployed independently, thus allowing faster responses to changes in the market and technology.

**Enhanced Security Features:** In this case using the blockchain technology allows better management of data and prevents data tampering where there are unauthorized users.

**Personalized Customer Experience:** This is again aided by the use of verdant AI which helps in sifting through customers' information and being able to tailor make services to them which enhances their satisfaction and loyalty.

#### 2.1.3. Comparative Analysis

Looking back, it is understandable that such centralised systems architecture worked in the development of few systems for airline reservations, but as the industry grew, so did the challenges associated with centralised systems, including scaling, inflexibility, poor security, and management. However, contemporary solutions have been developed to cope with these transformations by applying new technologies, which provide a better design, more effective, and more secure systems of managing such services as airline bookings.

### 2.2. Cloud Microservices in Airlines reservation

Due to the advantages of cloud microservices architecture in providing more scalability, flexibility, and efficiency for high demand applications, the aviation industry has employed the usage of this architecture greatly. This means that instead of a monolithic system being used, several systems or components are independently developed, deployed and scaled.

### 2.2.1. Adoption in Airline Reservation Systems

Most of the existing airline reservation systems are still very much monolithic and therefore suffer issues of scaling and maintenance. To solve these problems, cloud based microservices architecture have been suggested by many studies. It has been shown by Barua and Kaiser in a Cloud-Enabled Microservices Architecture for online airline reservation systems. Their solution is built on cloud computing with an incorporation of GDS, LCC, and Flight APIs that allow system enhancement in terms of scalability, fault tolerance, and efficiency. In the said system, performance tests conducted revealed that it could support up to 10,000 concurrent users at an average response time of 590 ms at peak loads.

### 2.2.2. Enhancements in Flight Management Systems

The principles of cloud microservices have also been introduced in flight management systems for increased operational effectiveness. Underwood et al. studied a model of a Cloud-Based Flight Management System (Cloud FMS) which proposes that the FMS of the aircraft is divided into two verticals [14]. Part of the FMS is integrated within the aircraft and the other part is off air in the cloud. This structure is expected to improve the flexibility and the reaction time of the system in relation to changes in flight for instance.

### 2.2.3. Infrastructure Inspection Using UAVs

Matlekovic and Schneider-Kamp (2022) explained the transition from monolithic architecture to microservices architecture to plan the infrastructure inspection procedures using an autonomous Unmanned Aerial Vehicle (UAV) [15]. They elaborated on better processing time and scalability that was brought about by changing the way the route calculation algorithm was done by breaking it into services and deploying it in a kubernetes cluster [16].

### 2.2.4. Performance Optimization

There have been efforts toward optimizing system structure performance due to increased adaptation of microservices in the cloud computing oriented systems. Zeng et al. reviewed the techniques used in microservice systems architecture to enhance system performance [23]. The authors observed the need to factor economies of resource utilisation and management, especially in high performance systems such as those in aviation.

### 2.2.5. Security Considerations

There are many challenges that should be solved in realization of the microservices oriented architecture and security is one of those. To secure geographical dispersed multi-domain avionics systems Xu et al. have described the Hybrid Blockchain Enabled Secure Microservices Fabric [18]. In this solution, the blockchain technology is applied to ensure data safety and security in different domains of aviation.

The use of cloud microservices architectures in the field of aviation applications was found to be effective in particular in terms of scalability, efficiency and security [19]. The research in progress also deals with the issues of the efficient performance of the systems and the safe terminal working with the data that drives the development of aviation industry.

## 2.3. Machine Learning and Artificial Intelligence in Reservation Systems

The application of Machine Learning (ML) and Artificial Intelligence (AI) in enhancing reservation systems has improved the predictive, prescriptive and automative functions of the systems and thus boosted the operational efficiency of the systems as well as that of customers [20].

### 2.3.1. Predictive analytics

With the use of historical booking data, customer behavior, and external parameters, algorithms AI and ML may also be employed to analyze and predict demand [21]. This creates a possibility for airlines to set better pricing controls and better management of the available seat inventory. For example, Sabre Dynamic Pricing system involves AI powered dynamic pricing that very frequently adjusts pricing based on forecasted demand enabling better revenues as well as occupancy levels [20].

### 2.3.2.Recommendation Systems

The need to develop loyalty schemes and services, AI Recommendation systems processes the consumers' preference data and their purchase history in order to recommend existing or alternatively travel products. It improves clients' satisfaction and encourages further spending. In case, such intelligent system would propose enhancing one's seat, what would one do with it assisting with additional offers for an already purchased ticket? Seating bonsai in economy class with in-flight services directed at final destination tourist attractions readily available on the airplane [22].

### 2.3.3.Automation and Operational Efficiency

The automation of reservation systems aided by artificial intelligence assists in easing some workloads from the systems and thus decreasing the levels of manual work and the corresponding costs incurred. Customer queries, bookings, and cancellations are done by chatbots and virtual assistants who operate all day long, allowing human agents to handle more difficult issues. It is also important to note that artificial intelligence systems can analyze and forecast the delays of flights and these assist airlines to manage their schedules, informing customers thus enhancing the quality of services delivered [24].

**Case Studies**

**Alaska Airlines:** The airlines began the use of the Flyways AI platform in the optimization of flight operations. This has facilitated proper routing and timely performance [25] [26].

**American Airlines:** The airline turned to AI and analytics as a means to reduce disturbances and improve service delivery, ultimately enhancing the passengers' experiences [28].

The modern implementation of AI and ML in reservation management systems has brought significant positive change in the aviation sector as it has made the business empirical, customer centric and less human [27]. The above positive outcomes in increase in finances, productivity levels and improving services to existing clients, point out the importance of AI and ML in contemporary reservation systems.

## 2.4. Blockchain for Data Integrity and Security in Airline Operations

In recent years, blockchain technology has turned out to be a game-changer in the improvement of data integrity and security across many sectors. Aviation has not been ignored. Its decentralized and tamper-proof ledger system ensures transparent data management and in a sector such as airlines characterized by complexity and interconnectivity, this is key.

### 2.4.1.Enhancing Data Integrity

In airline operations, constituents include but are not limited to ticketing, baggage, and maintenance, calling for accurate and consistent data. Blockchain technology comes in handy as its distributed database allows the input of certain information without the risk of modification or deletion on the existing information. This comes in equally handy when attempting to track the history of various parts used in maintaining an aircraft. It helps to ensure that parts used in constructing and maintaining an aircraft are safely used and possess fine documents (Brown and Wilson, 2023).

### 2.4.2.Improving Data Security

Establishing Control Measures to Increase Security of Data Because there is no central authority in a blockchain system, there is no single point of failure thus limits chances of hacking. All the transactions are coded using different encryption and added to a previous one to form a virtual chain making it impossible to allow breakout. In this respect, the security of information including sensitive details such as customer information and related payments in fulfilment of carriers' services is improved given the fact that the data will only be accessed and manipulated by authorized users [39].

### 2.4.3.Applications in Airline Operations

Management of Passengers' ID's: The process of validating a passenger's identity can be made easier by the use of blockchain providng a safe and authentic digital persona. This eliminates excessive checks and fastens the board time, which is an advantage to the passengers.

Supply Chain Tracing: Due to the ability of airlines to log and store every part and service transaction onto a software-controlled ledge, also known as a blockchain, it makes it possible for them to achieve bew greater transparency and traceability. This is to guarantee that all parts installed in an aircraft are genuine and fit for use as per the legal provisions hence improving safety and compliance [2].

Maintenance Records: Maintenance practices are reinforced with a blockchain system as it disallows the alteration of data with regards to maintenance practices done in an aircraft which also indicates that every practice is performed and can be verified when needed, which is especially important for reporting purposes and safety preservation of the aircraft [56].

### 2.4.4.Challenges and Considerations

Although the advantages of blockchain systems are extensive, there are hindrances to applied Blockchain technology in airlines operations such as merging with old technology, concerns about capacity, and the need for uniformity across the industry. Resolving these issues would require the convergence of the concerned parties and realistically assessing what the technology can do [47].

It is clear that the use of Blockchain technology could significantly enhance the security and integrity of the data processed in airline business. It could serve in improving interrelated operations including but not limited to maintaining and stocking spare parts, checking-in passengers and securing the transactional processes between the airlines [57]. As the upside in using the technology is apparent, there are also notable implementation challenges that must be put into serious consideration.

# 3. Research Methodology

## 3.1. Design of the Proposed System

The architecture of the system proposed comprises of several cloud microservices, artificial intelligence (AI) systems as well as the use of blockchain technology in order to solve the relevant operational challenges of the airline reservation systems. Such integration is purposeful in increasing the scale of operations, increase the degree of customer personalization and safeguard information.

### 3.1.1. System Architecture Overview

**Cloud Microservices as the Core Framework:** Cloud microservices facilitate the need for modularity, flexibility and scalability where each function of the reservation system such as booking, payment or customer management can function independently. This modularity enables the scaling up or down of system resources as per needs which is critical for real time high demand systems [37].

**Integration of AI Modules**: AI modules in the architecture help in predictive analytics towards demand forecasting, dynamic pricing and automation of customer care services [38]. These modules employ a number of machine learning algorithms and are aimed at enhancing decision making and improving customer services [39].

**Integration of AI Modules:** Blockchain technology to ensure transactions are made in an effective and contagion-free Manner. Such components ground the fact that all transactions are safe, clear as well as retained in a time capsule which is very protective of the customer's information including the financial records [31].

### 3.1.2. Cloud Microservices Architecture

**Microservices Design:** Every element of reservation system resides and deployed as individual microservices within a cloud environment to provide high availability and resiliency [29]. The primary services are:

**The Booking and Reservation Service:** This handles the real time reservations and connects with GDS (Global Distribution System).

**Payment Processing Service:** This administers the transactions with an inclusion of blockchain in the system to store the records of payments in an immutable and safe way [30].

**Customer Profile Management:** Preserves and evaluates customer profiles to ensure delivery of services that are tailored to the individual.

**Flight Management:** This ensures that real-time flight schedules are managed and any other changes are communicated to other services as required.

**Service Communication:** Microservices provide embedding mechanisms like REST APIs or gRPC for interactions among them enhancing the degree of effectiveness of the communication. Load balancers serve the traffic raised with reduced latency and down time [44].

### 3.1.3. AI Module Integration

**Demand Forecasting and Dynamic Pricing**: Artificial Intelligence models look at past and current data to assess demand and prices and make changes accordingly because it helps to increase earning [40] [41].

**Personalized Customer Service:** Such machine learning systems build a profile of users and offer them tailored content and services [42].

**AI-Powered Chatbots:** Chatboxes help in dealing with customers at any time of the day and assist a customer in making queries and even in booking. This is made even better with the help of Natural Language Processing as it magnifies the quality of the customer interaction [36].

**Operational Optimization:** Efficiency in operations has also significantly improved due to the application of AI techniques in resource allocation, maintenance scheduling, and also in predicting delays [43].

### 3.1.4. Blockchain Integration for Security and Transparency

**Decentralized Ledger for Transaction Security:** Blockchain as a database does not only keep records of a given transaction in the case of bookings, cancellations and even payments for services rendered or goods purchased but does so in an incorruptible manner to ensure [33].

**Passenger Identity Verification:** The use of check-in blockchain digital identities facilitates passenger check-in processes without compromising the user's identity [32].

**Smart Contracts for Automated Payments:** With embedded contracts, payments, cancellations, and reconciliation of loyalty points become less cumbersome because the processes become machine mediated rather than human [34] [35].

### 3.1.5. Integration Flow of Cloud, AI, and Blockchain

**Data Flow Management:** Cloud will act as the main storage site, encompassing micro services, AI, and blockchain activities among others. A centralized pipeline makes it possible for data to move smoothly.

.

**Illustration of A Working Scenario:**

- **Booking:** In the event that a passenger books a seat, the booking service communicates with the artificial intelligence modules for options before payment is recorded on the blockchain.
- **Customer Support Interaction:** Questions are fielded by AI chatbots who correct information using data checked on the blockchain.
- **Flight Delay Management:** A.I assisted predictive model forecasts those who are likely to be delayed — to compensate the passengers'. Changes, though, are managed by smart contract.

#### 3.1.6. Expected System Benefits

- **Scalability:** The implementation of cloud microservices ensures high degrees of growth potential while handling incoming traffic during peak seasons without much difficulty.
- **Enhanced Security:** Provided by blockchain is another protective mechanism that avails itself for use with the data.
- **Improved Customer Experience:** AI takes customer service experiences to a deeper level thanks to its ability to tailor services to individuals.

### 3.2. Data Collection:

In order to assess the system that is proposed, data will be gathered on various factors including performance, transaction speed and throughput, scalability, and security using a mixture of monitoring tools, synthetic testing, and security testing.

**Systems Performance:** Performance of the system in use is assessed using real-time monitoring tools such as Prometheus and Grafana that established CPU utilization, memory usage and response times [47]. In addition to that, there is JMeter which is a load testing tool that creates synthetic loads under high traffic conditions, in order to evaluate latency and error rates [49].

**Transaction Speed:** The different confirmation times are available on the transaction logs of the blockchain, while Zipkin tools help in measuring the service latency between various services [48].

**Scalability:** Auto-scaling and load-balancing metrics obtained from cloud-based services, for example, AWS ELB determine how much the system can withstand increasing loads, its elasticity [45] [46].

**Security:** Vulnerability assessments check how safe a blockchain is, whereas every microservice is tested by way of penetration comprehension using OWASP ZAP [11] [50].

These methods consider this system's demand, transaction security, and resource management capabilities exhaustively.

### 3.3. Analysis Techniques

The proposed system will be evaluated by carrying out a comparative analysis, simulations, and machine learning assessments.

**Comparative Analysis with Traditional Systems:** In order to create the benchmarks and estimate the gains, key performance metrics (KPMs) such as latency, transaction throughput, and scalability of the existing airline reservation systems would be used [12] [51].

**Simulation:** Moreover, AnyLogic and MATLAB will be deployed in simulating high demand and peak traffic circumstances. This will enable testing of the system's scalability, robustness and faults tolerance capacity under controlled conditions, particularly with respect to load variations typical in airline booking systems [13] [52].

**Machine Learning Techniques for Evaluation:** Data coming in will first be predictive modeled so as to allow understanding when the resources would be poorly managed due to bottlenecks [53]. Regression and anomaly detection algorithms will prove useful in understanding how the system scales and responds to different loading conditions [54].

These analysis techniques provide a holistic approach to validate the system and show areas which may require improvements.

## 4. System Architecture Design

The system architecture is envisioned as a modular, cloud-based microservices architecture to offer scalability, availability, and service management independently [55]. Each service of the architecture – Reservations, Payments, and Customer Data – is autonomous as it performs different roles and communicates through clear-cut APIs. It is possible to adjust only those services that are experiencing high demand while the other services remain unchanged, which is very important in the ever-changing airline sector. Furthermore, the system incorporates artificial intelligence-based modules for advanced statistical analysis and customer satisfaction enhancement as well as blockchain technology for increased data security and clarity of transactions. This kind of architecture enables very efficient and secure operations, such that each service is balanced to support high speed and quality of work while delivering good service with minimal disruption to the users.

### 4.1. Cloud Microservices Architecture: Detailed Structure and Design

The cloud-based microservice architecture transforms the airline reservation system into several cohesive, self-sufficient deployable units, namely reservations, payments, and customer information management. Every microservice is designed to focus on scalability, fault tolerance, and independent of each other update.

#### 4.1.1. Structure and Design of Microservices

- **Reservation Service:** Provides support for booking, availability, and itinerary related operations.
- **Payment Service:** Responsible for carrying out transactions, making payments and checking payment status.
- **Customer Data Service:** Responsible for storing and looking after customer's profile, preferences and loyalty details.

#### 4.1.2. Algorithm for Booking Process

Algorithm BookFlight(customer_id, flight_id, payment_details):

1. Check availability of flight_id in Reservation Service
2. If available:
    a. Create a booking entry using the flight and customer IDs.
    b. Send booking ID to Payment Service
    c. Process payment with payment_details
    d. If payment successful:
        i. Confirm booking in Reservation Service
        ii. Update customer loyalty points in Customer Data Service
        iii. Send confirmation to customer
    e. Else:
        i. Cancel booking in Reservation Service
        ii. Notify customer of payment failure
3. Else:
    a. Notify customer of unavailability

#### 4.1.3. Booking Process

The procedure for making a reservation commences with the inquiry of seat availability for the requested flight. Where seats exist, a temporary reservation is made, and a request for payment is forwarded to the Payment Service for action. After this payment is completed, the booking is made, and the loyal customer's points are adjusted appropriately. In case of a payment glitch, the booking is voided automatically, and the customer is apprised. If all the seats are already filled, the customer is updated without delay. This has enabled a fast tracking of the process for the benefit of the customer with every important stage implemented and updated automatically. The flowchart has been shown in figure 1.

**Flowchart**

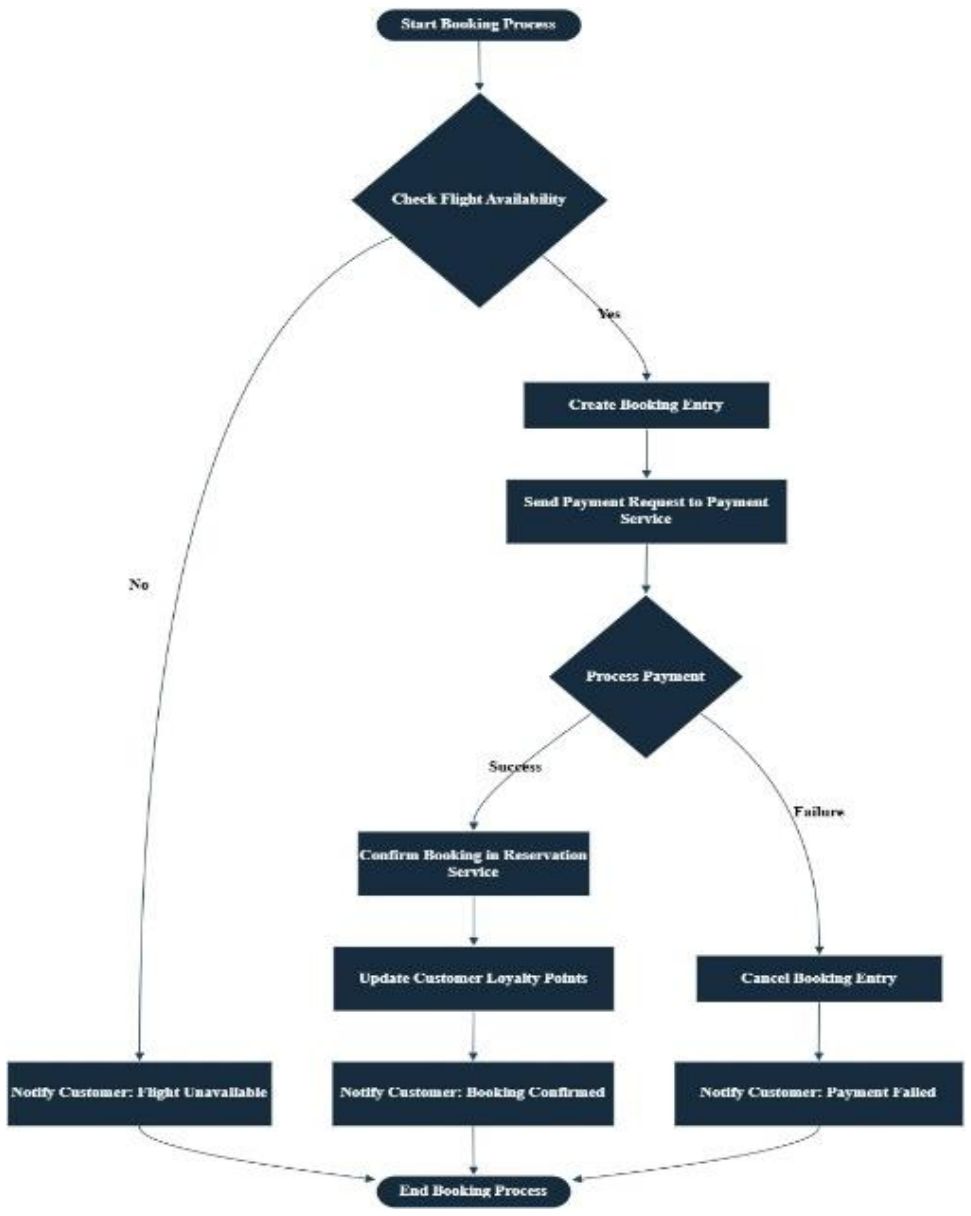

**Fig.1.** A flowchart of flight availability checks to payment confirmation and customer notification

### 4.1.4. System Diagram

The system diagram shows how the microservices interact more specifically with the airline reservation architecture. In the figure 2 it shows how Reservation, Payment, and Customer Data services work through clean APIs to support the processes of booking, payment, and customer management. The diagram also depicts the information flow from the customer request to the booking confirmation and the cloud microservices, AI, and blockchain's integration in making the whole process reliable, available, and secure. Such clear illustrations are useful as far as comprehending the data relationships and movement in the system is concerned.

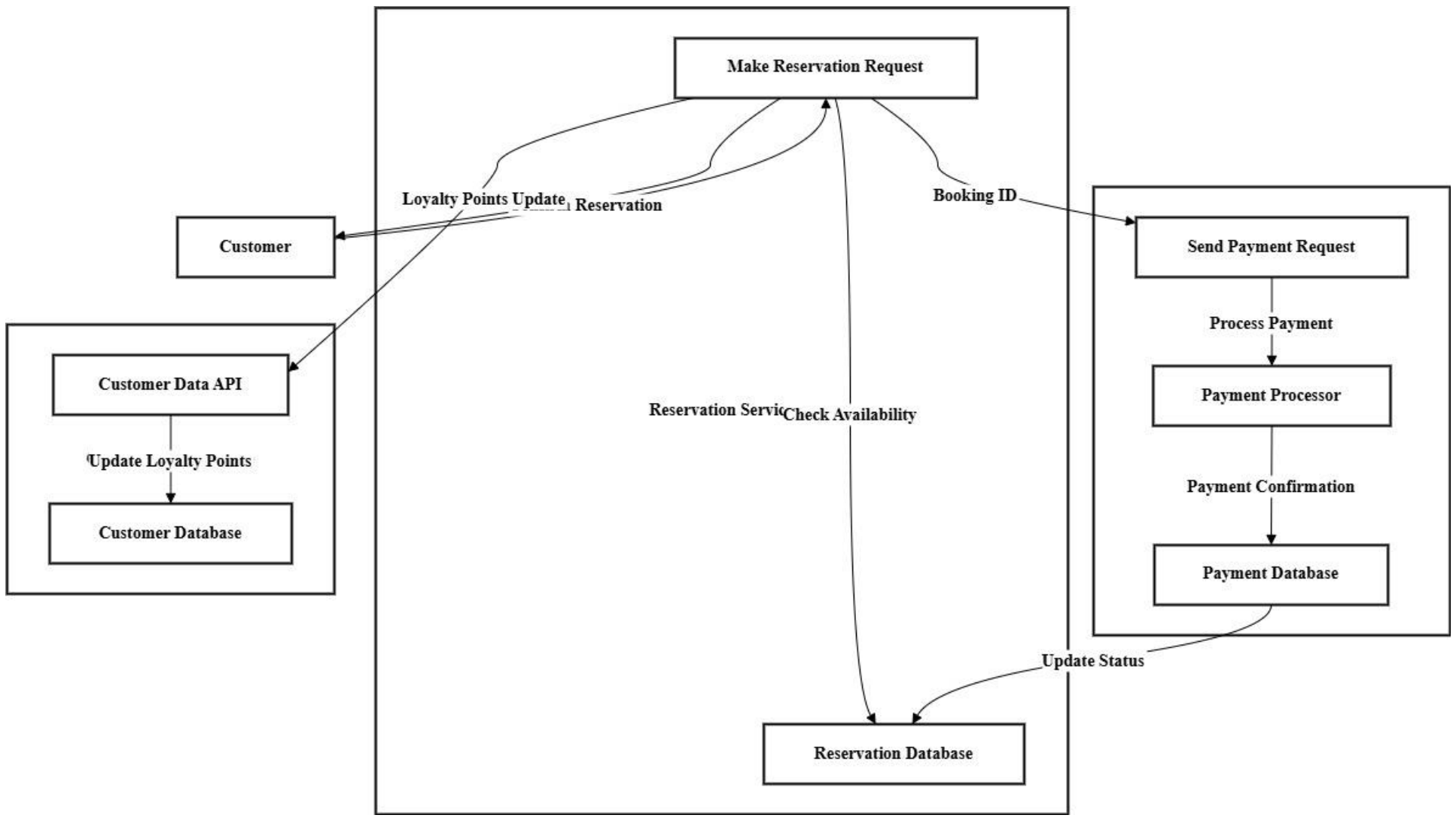

**Fig.2.** A diagram of flow between these microservices.

## 4.2. AI-Driven Modules

The proposed system's architecture is designed with two main aspects in mind that can be aided by the modules powered by AI technologies: demand forecasting and customer engagement tailoring. While prediction models assist the airline in planning its resources based on the likely sales, the personalization algorithms change the interaction with the customer in accordance with the differences between the individuals, enhancing the overall satisfaction of that person.

### 4.2.1. Prediction Models for Demand

The demand prediction model relies on historical booking records, time of year, holidays, and other external influences in order to project passenger demand accurately. This allows the airline to make pricing changes, allocate planes, and ramp up for peak demand. The data structure shown in table 1 below is used in the prediction model:

**Demand Data Table**

**Table 1.** Demand data table.

| Date | Flight_ID | Day_of_Week | Holiday_Flag | Previous_Demand | Forecast_Demand |
|---|---|---|---|---|---|
| 2024-11-09 | 101 | Saturday | No | 180 | 200 |
| 2024-11-10 | 102 | Sunday | Yes | 250 | 300 |

#### 4.2.1.1. Algorithm for Airline Demand Prediction

Airline reservation system requirements can easily be predicted with the help of the following step by step algorithm.

**Stage I Data Collection.**

This stage entails collecting embedded data regarding historical booking records such as travel date and destination, airlines, and available seats, and fare classes. Furthermore, internal perishable data such as public holidays, special programs, competitive fares and economy facts like GDP growth must be obtained.

**Stage II Data Preprocessing.**

At this stage, the data is cleaned and made ready for analysis through transformation to take care of missing values and outlier points. Variables like routes, corresponding dates and fare classes will be standardized for uniformity.

**Stage III Feature Engineering.**

Generate features related to the day of the week, season, holidays and events, fare class etc. Incorporate lags for prior observations of the amount of bookings along with moving ranges of observed book growth to account for the trend in current bookings.

**Stage IV Model Training.**

Employ relevant machine learning algorithms with time series data capability, for instance, linear regression, ARIMA or LSTM, to model the airline data.

**IV Stage Forecasting.**

Estimate how many individuals will make a reservation for each route, fare class, and date I have trained the models on.

**Stage V Evaluation.**

Check model evaluation metrics to see how well the model's predictions compare with actual outcome data.

Refine the model's parameters if the accuracy of prediction is not satisfactory.

#### 4.2.1.2. Sample Python Code for Airline Demand Prediction Using Linear Regression

In this section, a possible implementation is illustrated for a Python software on demand forecasting along specific routes and dates through an airline booking system with the help of scikit-learn tools.

```
import pandas as pd
```

```
from sklearn.model_selection import train_test_split
from sklearn.linear_model import LinearRegression
from sklearn.metrics import mean_absolute_error

# Sample data for airline demand forecasting
data = pd.DataFrame({
    'Date': pd.date_range(start='2024-01-01', periods=365),
    'Route': ['NYC-LAX'] * 365,
    'Day_of_Week': pd.date_range(start='2024-01-01', periods=365).dayofweek,
    'Holiday_Flag': [1 if date.weekday() == 5 else 0 for date in pd.date_range(start='2024-01-01', periods=365)],
    'Previous_Demand': [200 + (i % 20) * 10 for i in range(365)],  # Mock demand pattern
    'Competitor_Price': [300 + (i % 15) * 5 for i in range(365)]  # Mock competitor pricing pattern
})

# Target variable - shifting previous demand to simulate future demand
data['Forecast_Demand'] = data['Previous_Demand'].shift(-1)
data.dropna(inplace=True)

# Features and Target
X = data[['Day_of_Week', 'Holiday_Flag', 'Previous_Demand', 'Competitor_Price']]
y = data['Forecast_Demand']

# Train-test split
X_train, X_test, y_train, y_test = train_test_split(X, y, test_size=0.2, random_state=42)

# Model training
model = LinearRegression()
model.fit(X_train, y_train)

# Predictions and evaluation
y_pred = model.predict(X_test)
mae = mean_absolute_error(y_test, y_pred)

print("Mean Absolute Error:", mae)
```

```
print("Predicted Demand for Next Flight:", model.predict([[6, 1, 210, 320]]))  # Example prediction
```

In this case, we add a feature on the prices of competitors, as competition might affect the amount of demand. It is important to note that the model may be improved through the inclusion of other variables important in the airline business.

#### 4.2.1.3. System Flow Diagram

In this diagram in figure 3, we outline the process of demand prediction which is oriented to "airline reservation system".

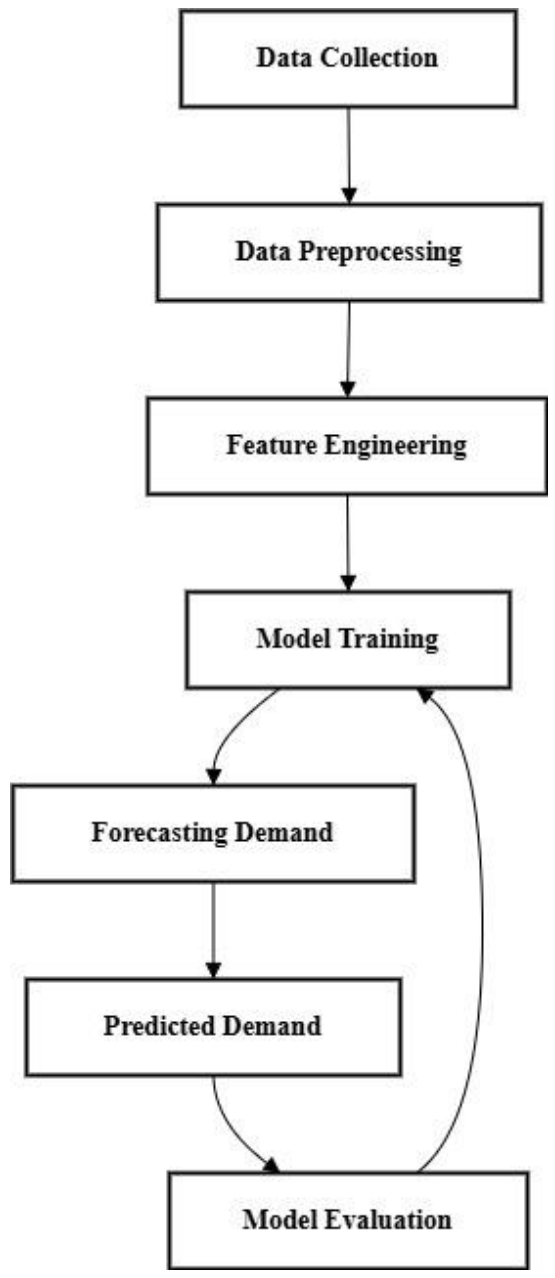

**Fig.3.** A diagram of demand prediction process specific to the airline reservation.

**Diagram Explanation**

**Data Collection:** Gathers information about travel, vacations, fare types and more, including the competition pricing.

**Data Preprocessing:** Treats and standardizes data for further analysis.

**Feature Engineering:** Develops airline's specific demand features (e.g. day of the week, holiday activity flags, fare type).

**Model Training:** With the using the prepared data, trains designed predictive computational model where the application relates to forecasting demand for air carrying business.

**Forecasting:** Makes a demand forecast on a particular route for a given date and other specified features.

**Model Evaluation:** Assesses how accurate the forecasts generated by the model are and updates the model if need be to obtain better predictions.

This well-defined procedure of demand prediction incorporated in the airline reservation system allows proper management of seat allocations and pricing policies to the airline's offer in order to respond to periodic changes in demand. I can help with more personalization if needed!

### 4.2.2. Personalization Algorithms for Customer Preferences

Cognitive computing personalizes user experiences by looking at customer behavior like previous bookings, seat preference, and habitual places of call. After this evaluation, the algorithm recommends a flight and a seat, as well as additional services. Such an approach enhances the customer experience and retention in the business. The data are shown in table 2.

**Table 2.** Customer preference data

| Customer_ID | Preferred_Seat | Frequent_Destination | Past_Bookings | Loyalty_Points |
|---|---|---|---|---|
| 1001 | Window | New York | 5 | 1200 |
| 1002 | Aisle | London | 8 | 1800 |

#### 4.2.2.1. Algorithm for Personalization Based on Customer Preferences

**Data Collection:**

- Collect customer data such as previous entries of booking, choice of seats, commonly visited places, the loyalty points, and buying habits.
- Collect understanding or appreciating information, such as, number of travel trips in a fiscal year, particular season preferred and days of the week for travel.

**Data Preprocessing:**

- Process women customer's data so as to eliminate missing values and make the data consistent.
- Standardize the purchase history and preference in order to eliminate bias in the created profile.

**Feature Engineering:**

- Introduce feature like preferred seat, aisle or window, common travel destinations, loyalty tier, and mean advance purchase.
- Create new features in reference to existing feature, for example 'Customers who spend a lot' or 'Travelers who travel during weekends' for better profiling.

**Personalization Model:**

- Approaches like collaborative filtering or content based filtering techniques to make recommendations to users are implemented
- Using models such as decision trees or neural networks, learn customer patterns and make appropriate recommendations based on the learned customer preference.

**Recommendation Generation:**

- Depending on the profile of the customer recommend flights, seats, and services.
- Change recommendations made to the customer depending on the preferences and behavior of the customer at that time.

**Evaluation and Feedback:**

- Monitor or log the users of the recommendations and how may of that helps improve the recommendation's system.
- Modify recommendation quality based on consumers' reaction and contentedness measure.

#### 4.2.2.2. Example of a Personalization Algorithm

Collaborative filtering algorithms can be employed in an airline system. Below is a simple pseudocode for such personalization algorithm:

Algorithm GenerateRecommendations(customer_id):

1. Retrieve customer preferences from past bookings and purchases

2. Find similar customers using collaborative filtering based on:

a. Frequent destinations

b. Preferred seat type

c. Loyalty points or membership tier

3. Create tailored suggestions by analyzing typical behaviors:

a. Suggested routes or flight times

b. Preferred seat options (e.g., window, aisle)

c. Recommended add-ons (e.g., lounge access, extra baggage)

4. Output recommendations to the user interface for customer selection

5. Record feedback to improve future recommendations

#### 4.2.2.3. System Flow Diagram

The following diagram illustrates how the personalization algorithm interacts with other components of the airline reservation system to generate tailored recommendations.

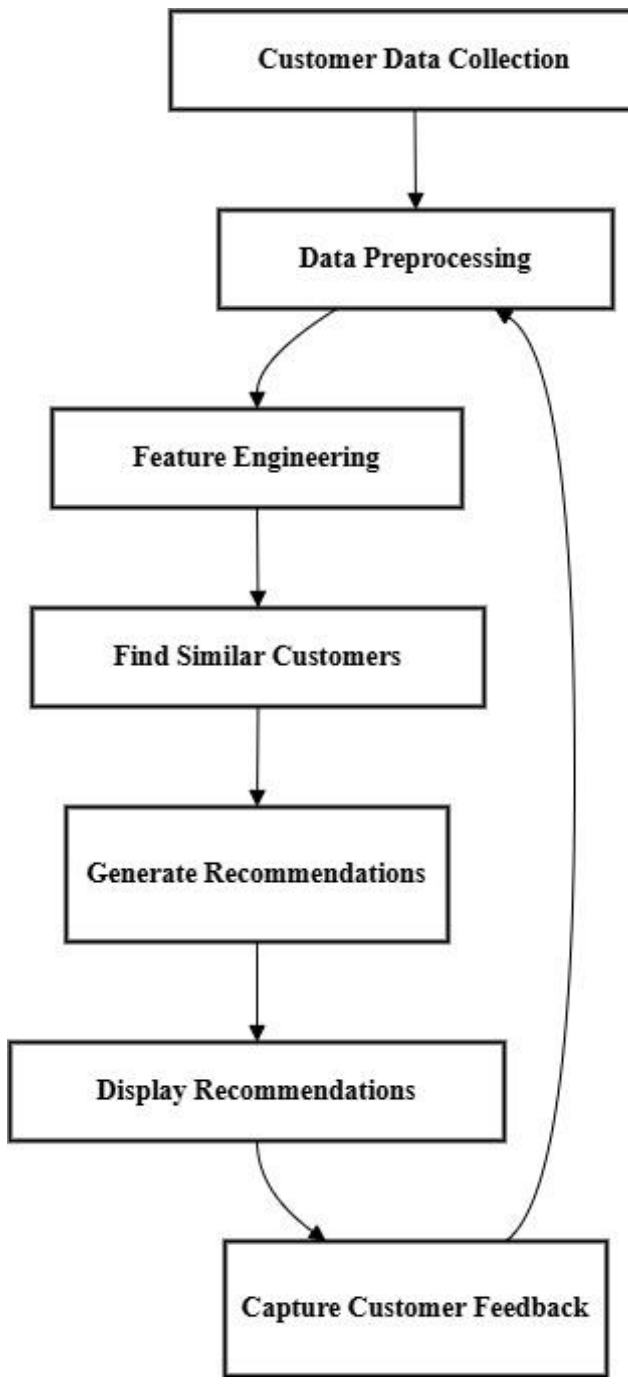

**Fig.4.** A diagram illustrates how the personalization algorithm interacts.

Such architecture in figure 4 offers an effective solution to creating appropriate suggestions in relation to a customer's interests, fostering engagement and appreciation of the brand's services.

### 4.2.3. System Diagram for AI-Driven Modules

The system diagram of AI based modules in which demand forecasting and customer personalization algorithms are shown to be fitted within the airline reservation structure. These Modules incorporate past Booking records, customers' preferences and preferences, as well as external environments to make smart decisions and suggestions. The demand forecasting module calculates possible future bookings per route and period in order to optimize the resources allocation, while a personalization module provides specific recommendations on flights, seats, and other amenities based on a given customer's preferences.

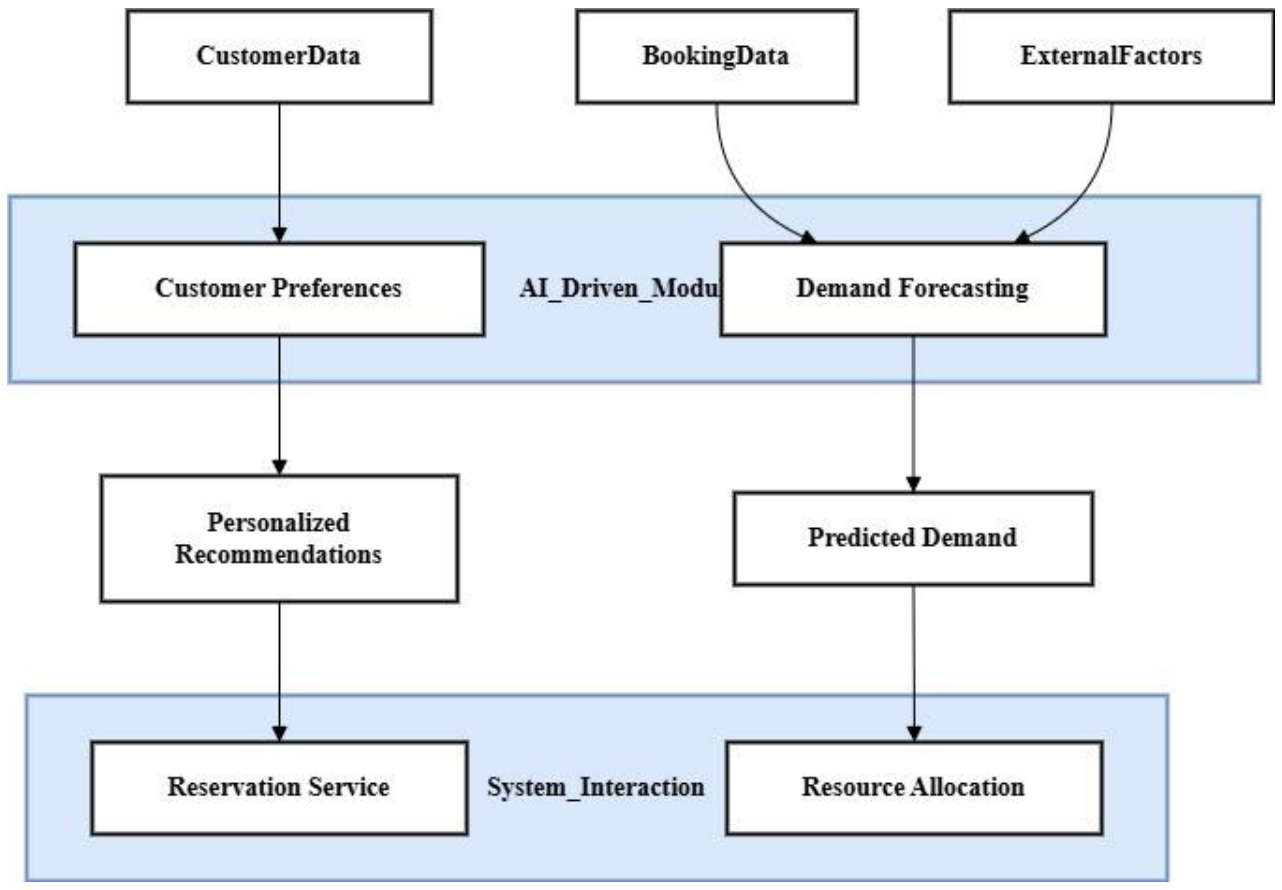

**Fig.5.** Diagram illustrates how the AI-driven modules for demand prediction and personalization.

By integrating the figure 5 above described AI modules with the reservation system, the airline is able to offer an adaptable and customized purchasing process which promotes the efficiency of operations while at the same time boosting the satisfaction of the clients.

### 4.3. Blockchain for Security and Transparency

Incorporating blockchain technology into the airline reservation system improves both security and transparency. Thanks to the decentralized and censorship resistant nature of the blockchain, customers are ensured complete fairness in dealing with the airline, preventing cases of data loss or even theft. Best examples of such transactions are recording users' transactions history, confirming users and their transactions and monitoring their engagement in or earning of rewards programs.

#### 4.3.1. Algorithm for Security of Transactions and Customer Data

In order to process secure transactions and handle customer information the following steps are envisioned for the system development:

**Data Collection and Data Encryption:**

System gathers information about each transaction, customers and their loyalty scores.

System encrypts any private information before entering it into the blockchain and makes an effort to secure the confidentiality of the stored information.

**Transaction Initiation:**

Each and every transaction (e.g., on booking, on payment) has a transaction ID assigned and is formatted as a transaction for blockchain processing.

**Consensus Verification:**

A number of the consensus methods such as Proof of Work or Proof of Stake enable the network nodes to validate the transaction by consensus for compliance to network rules.

**Transaction Recording:**

In the next stage, when the transaction is authenticated, it is included in the new block and attached to the existing blockchain creating a permanent record.

**Customer Data Retrieval:**

Customer data is obtained from the blockchain if required, for example, during loyalty points verification, ensuring no loss of data.

**Updating Points:**

When customers perform specific actions like bookings, their loyalty points which are kept on the blockchain are updated instantaneously.

**Audit and Compliance:**

To guarantee integrity and conformity, periodically audits are done by pulling out transaction history from the blockchain.

### 4.3.2. Sample Pseudocode for Blockchain Transaction Process

Algorithm ProcessTransaction(customer_id, booking_id, transaction_details):

1. Encrypt customer_id and transaction_details for security
2. Generate transaction_id and create blockchain transaction
3. Initiate consensus mechanism:
   a. Verify transaction through network nodes
4. If transaction is verified:
   a. Update the blockchain ledger with a transaction.
   b. Update customer loyalty points if applicable
5. Notify customer of successful transaction
6. If transaction fails verification:
   a. Discard transaction and alert customer of failure

### 4.3.3. System Flow Diagram

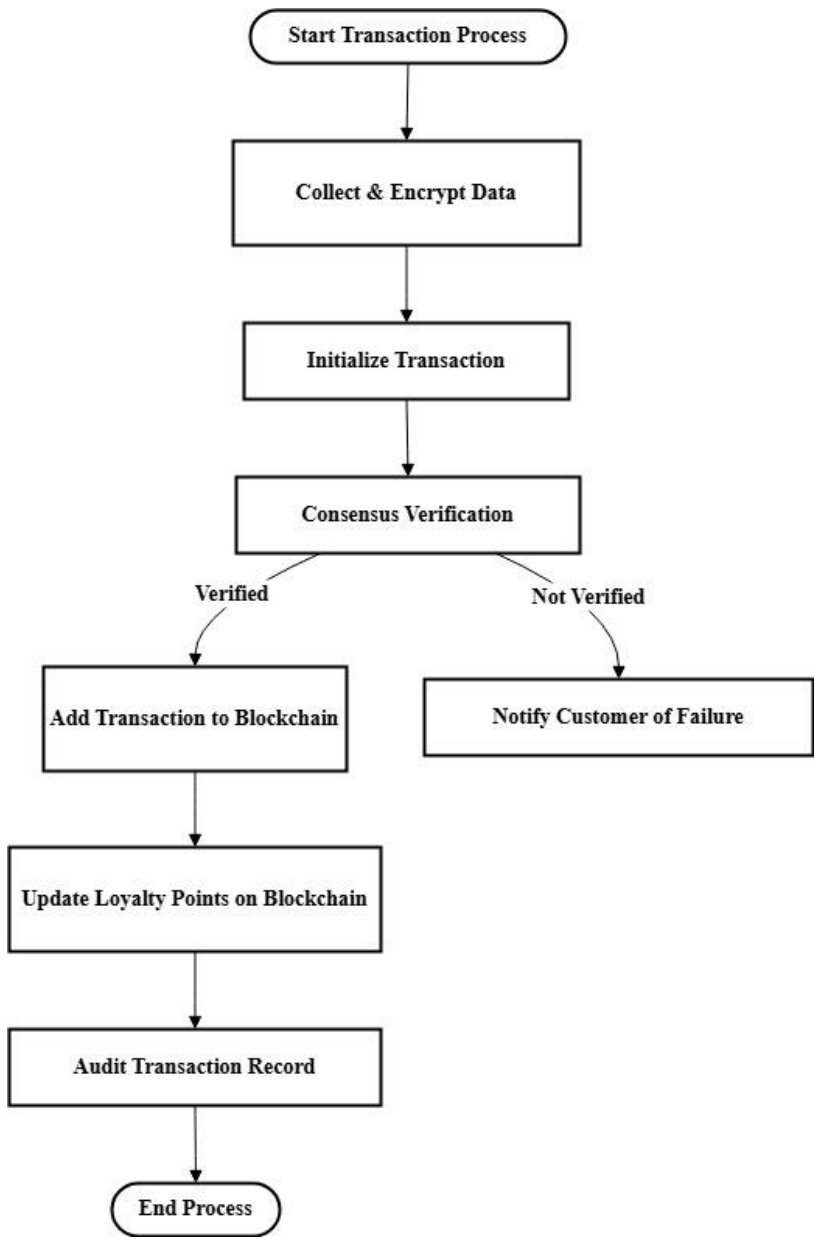

**Fig.6.** the blockchain process for handling secure transactions and managing customer data.

The details diagram of data flow is shown in figure 6.

**Description**

1. **Start Transaction Process:** Handles the initiation of the transaction process and data management in a secured way.
2. **Data Collection & Encryption:** In this stage customer details and transaction detail is gathered and encoded with secrecy.
3. **Initialize transaction:** A unique id of the transaction is created and this id is stored in the blockchain.
4. **Consensus Verification:** Validates the transaction on the network by consensus. As a result of that, the transaction moves on to the next stage, if verification is done. Otherwise, the client gets an alert that there was a failure.
5. **Add Transaction to Blockchain**: This process entails updating the transaction in the blockchain ledger where it becomes permanent.
6. **Updating of Loyalty Points on Blockchain:** The update of the loyalty points is done automatically for the customer since it is stored in the system as an added advantage.
7. **Audit Transaction Record:** Provides the means of systemically reviewing transaction records for the purposes of security and adherence to regulations.

The integrity of the data and tribal trust is enhanced as this blockchain structure facilitates the proper and accountable management of airline transaction and customer information.

### 4.4. Integration Flow: How Cloud, AI, and Blockchain Interact within the Reservation System

In the current airline reservation system design, cloud microservices, artificial intelligence systems and blockchain technology work together to secure, streamline and bring customization to the travel booking experience. The cloud contains and operates scalable microservices, such as: Reservation Service, Payment Service, Customer Service etc; AI Systems build customer demand forecasting and suggestive selling; while Blockchain promotes safety and reliability in transactions. In general, the aforementioned elements help improve the performance of airlines as well as to please customers more.

### 4.4.1. Data Flow Across the System

1. **Customer Data Acquisition:**
   In this case customer contacts, reservation records, and customs preferences are created and saved in the cloud based data storage.
2. **Demand Prediction and Personalization:**
   AI systems research past reservation patterns to predict future demand and help create unique suggestions for each user.
3. **Management Of Secure Transactions And Files:**
   Transactions and loyalty details are encrypted into teaching records through Blockchain making sure no data is lost and further all information is evident.
4. **Interaction Flow:**
   The customer makes a new booking, cloud microservices control the process and send the request to AI systems in order to get the suggestions.

**Example Data Tables:** The table 3 and table 4 are shown the customer preference and booking transaction details.

**Table 3.** Customer preference table.

| Customer_ID | Preferred_Seat | Frequent_Routes | Loyalty_Status |
|---|---|---|---|
| 1001 | Window | NYC-LAX | Gold |
| 1002 | Aisle | LHR-DXB | Silver |

**Table 3.** Booking transaction table.

| Booking_ID | Customer_ID | Flight_ID | Transaction_Status | Blockchain_Hash |
|---|---|---|---|---|
| B101 | 1001 | F101 | Confirmed | 0x1a2b3c... |
| B102 | 1002 | F102 | Pending | 0x4d5e6f... |

**Flowchart**

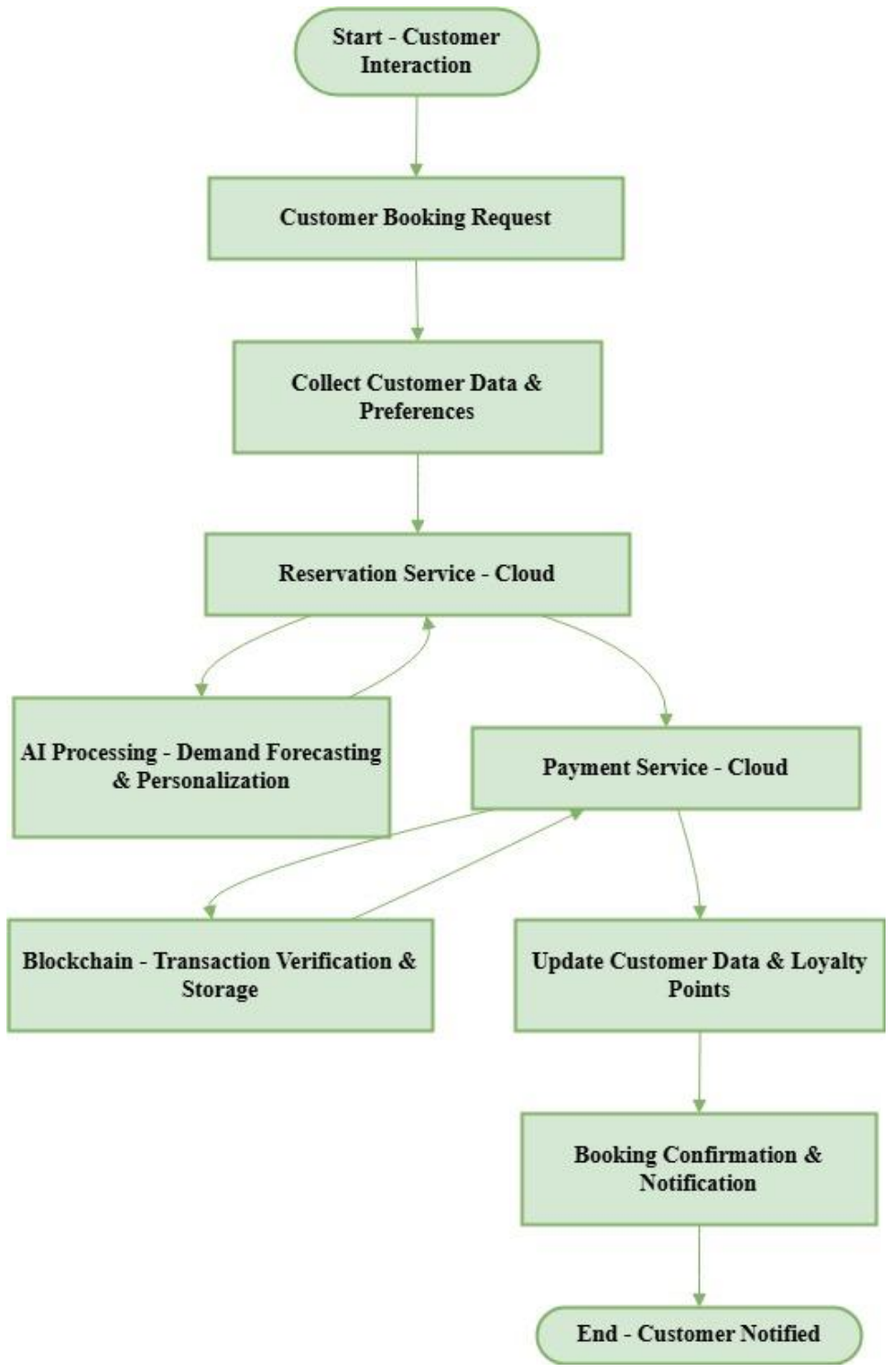

**Fig.7.** how data moves through the cloud, AI, and blockchain components

Every payment transaction is checked and encoded in the blockchain system for security as well as transparency purposes.

The integrated flow in figure 7 makes it possible to update customer information, loyalty point and booking status proactively without any hitch.

#### 4.4.2. Integration Flowchart

The description given below in figure 8 is a flow chart for the interaction of cloud, AI, and Blockchain components in an Airline Reservation System.

1. **Booking Process:** The customer starts OR initiates a booking request in the cloud bookshelf.
2. **AI Modules:** Produce demand forecasts and tailored suggestions.
3. **Reservation confirmation:** Blockchain authenticates the payment and processes the transaction.
4. **Customer Update:** Realtime escalations of the customer data, loyalty level, and booking confirmation.

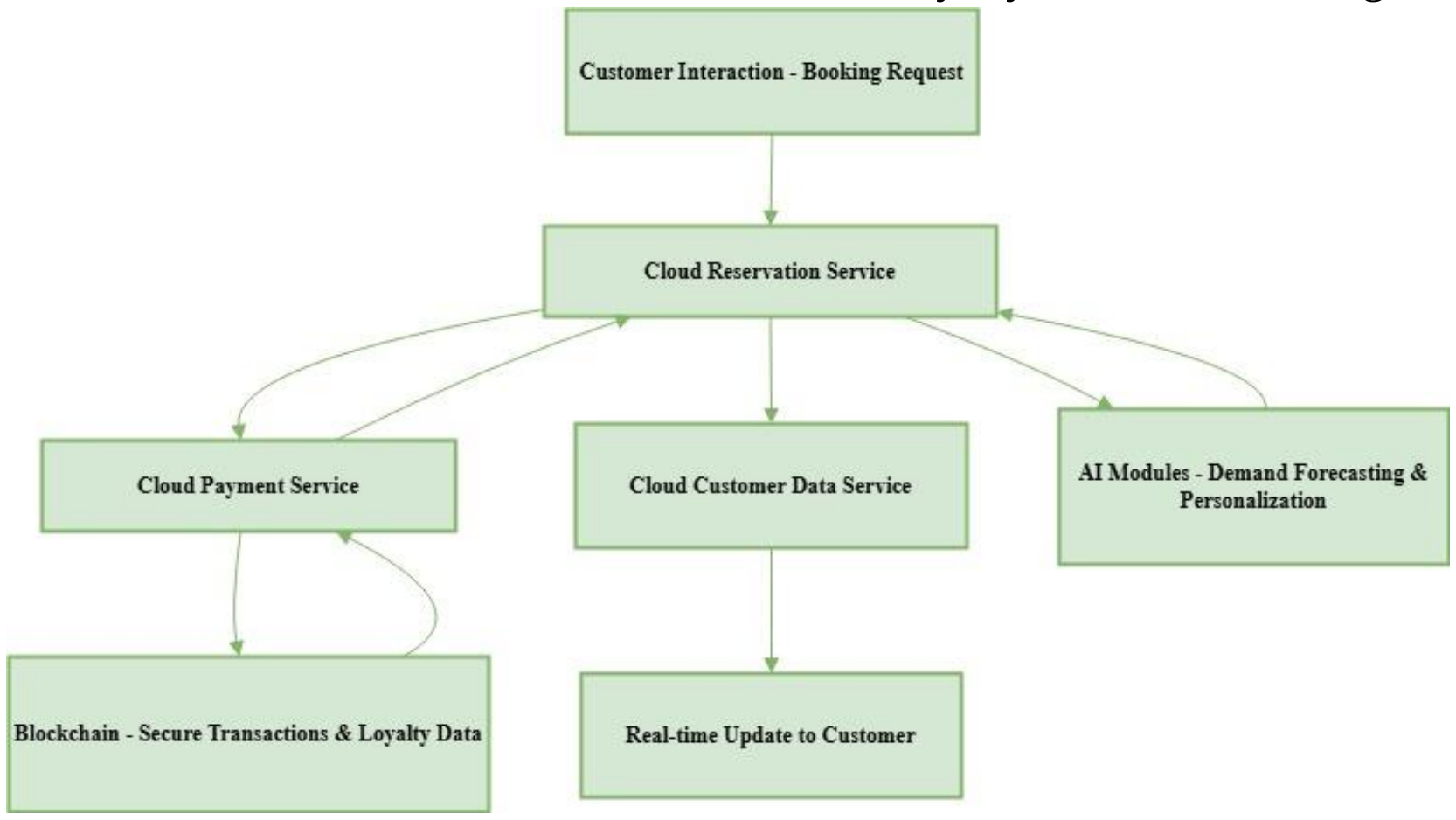

**Fig.8.** the integration flow among cloud, AI, and blockchain

## 5. Expected Benefits and Challenges

### 5.1. Advantages

**Improvement in the system:** If the system is developed using a cloud microservices architecture, resources will be managed effectively and every component of the system will work at its best. This approach also minimizes latency by realizing a modular structure making it possible even in high traffic reservation applications to process requests without delays in sending responses.

**Less System Downtime:** The cloud-based design allows for service maintenance or enhancements to be carried out at different times without taking the entire system off line. This architecture enhances service delivery and guarantees provision of services at all times even in times of rushes or in situations of unexpected increased demand.

**Transaction Safety:** The incorporation of blockchain provides a solution for the problem associated with a single point of failure of record keeping system as all transactions are recorded on an infinite chain of records. This environment of carrying out commercial transactions online promotes customer's confidence due to the protection of their private information and the openness of the system.

**Better Experience for Customers:** Booking modules enhanced with artificial intelligence help to improve the booking process by making personalized suggestions according to the user's preferences and previous booking history. Together with the great offer, it also satisfies their desire for a specific level of service, thus attracting them to the airline.

**Flexibility:** System can scale up and down, out and in to manage the cloud resources as required. Because microservices are scaled up or down independently from other parts of the system, it is also easier to manage growth in usage or functionalities for which the system is built for as the airline progresses.

### 5.2. Challenges

**Complex Integration:** To claim your IP over a given software app you may need to include all intermediate stages of creation in case of app development. Thus, it will make the process of deployment quite difficult.

**Costs:** To set up an integrated or all in one cloud and ai managed deployment platform will require quite a huge sum of money.

**Data privacy:** Keeping customers' details on the blockchain as well as abiding by the laws on privacy is hard.

System Maintenance: The system is built in components and for installation and maintenance of these components special skills are required which increase the complexity of operations.

**Delays in Performance:** With regards to performance optimization, care must be taken so as to ensure that all the parts work in harmony and there is no lag between them which would cause a slowdown of the system.

## 6. Future Considerations

The deployment of cloud microservices, artificial intelligence, and blockchain technology in airline booking systems can potentially change the airline business for the better, making it more efficient, secure and customer friendly. This site-specific, data intensive system might also be applicable to other industries, for example, the hospitality sector, transportation supply chain and the retail industry, where there are many secure and unambiguous payments and intense customer handling. Further investigations could look at additional advances in this area such as the application of the quantum computing and the Internet of Things (IoT) advancements in improved reservation systems. Quantum computing could handle better the complex processes of scheduling and route construction and in the case of IoT, it could introduce real-time updates improving the continuity of service to the client so as to stretch even further the limits of innovations in the airline.

## 7. Conclusion

To summary, this paper explores the benefits of implementing the airline reservation system using cloud microservices and AI technology with blockchain which guarantees performance, security, scalability and improves customer satisfaction. Cloud based microservices enable reliability and flexibility, Artificial Intelligence driven modules assist in predicting demand and personalizing services while Blockchain guarantee secure and transparent transactions. These technologies come together to bring about a hassle free, effective, and efficient reservation process.

The envisaged architecture will enhance efficiency and effectiveness, but it goes beyond that in terms of benefits. The adoption of emerging technologies enables the airline sector not only to redefine efficiency and focus on customers but also to achieve even more futuristic objectives like quantum computing and the internet of things. There is much indication that this change is a welcome development towards the development of reservation systems that are not only intelligent but are also able to adjust to changing scenarios and reinstate the expectations of the customers and practices of businesses in the concerned industries.